\documentclass[12pt]{spieman}  
\usepackage{amsmath,amsfonts,amssymb}
\usepackage{graphicx}
\usepackage{setspace}
\usepackage{tocloft}
\usepackage{booktabs} 
\usepackage{soul}

\title{Geometry-Aware Video Inpainting for Joint Headset Occlusion Removal and Face Reconstruction in Social XR}

\author[a,*]{Fatemeh Ghorbani Lohesara}
\author[b]{Karen Eguiazarian}
\author[c]{Sebastian Knorr}
\affil[a]{Technische Universität Berlin, Communication Systems Group, Department of Telecommunication Systems, Einsteinufer 17, Berlin, Germany, 10587}
\affil[b]{Tampere University, Computational Imaging Group, Department of Computing Sciences, Korkeakoulunkatu 10, Tampere, Finland, 33720}
\affil[c]{HTW Berlin, Visual Computing Group,Faculty of Informatics, Communication and Economics, Wilhelminenhofstraße 75A, Berlin, Germany, 12459}

\cftpagenumbersoff{figure}
\cftpagenumbersoff{table} 
\begin{document} 
\maketitle

\begin{abstract}
Head-mounted displays (HMDs) are essential for experiencing extended reality (XR) environments and observing virtual content. However, they obscure the upper part of the user's face, complicating external video recording and significantly impacting social XR applications such as teleconferencing, where facial expressions and eye gaze details are crucial for creating an immersive experience.

This study introduces a geometry-aware learning-based framework to jointly remove HMD occlusions and reconstruct complete 3D facial geometry from RGB frames captured from a single viewpoint. The method integrates a GAN-based video inpainting network, guided by dense facial landmarks and a single occlusion-free reference frame, to restore missing facial regions while preserving identity. Subsequently, a SynergyNet-based module regresses 3D Morphable Model (3DMM) parameters from the inpainted frames, enabling accurate 3D face reconstruction. Dense landmark optimization is incorporated throughout the pipeline to improve both the inpainting quality and the fidelity of the recovered geometry.

Experimental results demonstrate that the proposed framework can successfully remove HMDs from RGB facial videos while maintaining facial identity and realism, producing photorealistic 3D face geometry outputs. Ablation studies further show that the framework remains robust across different landmark densities, with only minor quality degradation under sparse landmark configurations.
\end{abstract}


{\noindent \footnotesize\textbf{*}Fatemeh Ghorbani Lohesara,  \linkable{ghorbani.lohesara@tu-berlin.de} }

\begin{spacing}{2}   

\section{Introduction}
\label{sec:intro}

\begin{figure}
\begin{center}
  \includegraphics[width=0.7\linewidth]{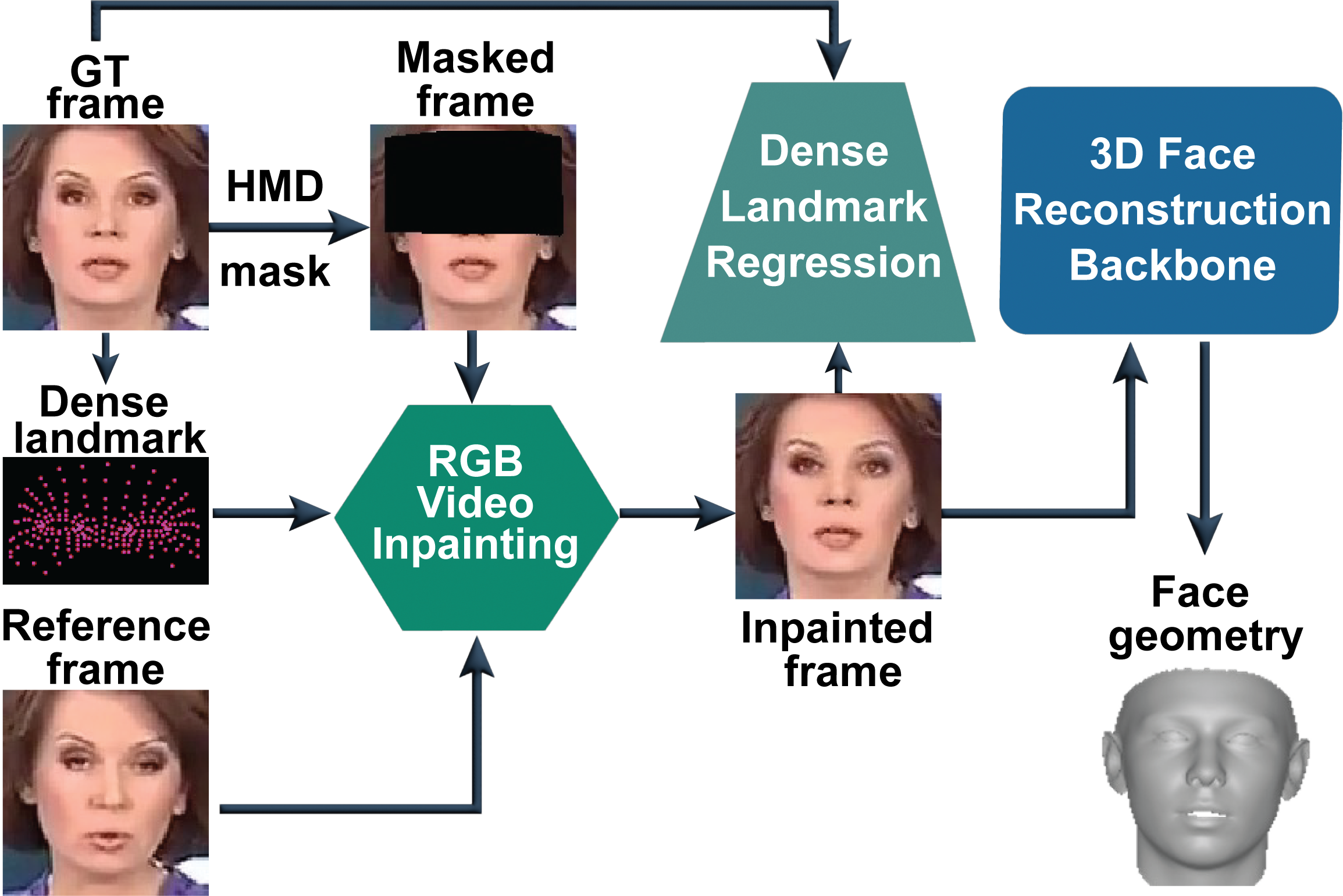} 
  \caption{Overview of our proposed framework for HMD removal and 3D face geometry reconstruction in social XR. The pipeline integrates RGB video inpainting, dense landmark regression, and 3D face geometry reconstruction to restore occluded facial regions. (images: TCH (\url{https://www.youtube.com/watch?v=2su-BnPJ9iY})).}
  \label{fig:teaser}
\end{center}
\end{figure} 

The rapid advancement of extended reality (XR) technologies has led to sophisticated head-mounted displays (HMDs) such as Meta Quest, Apple Vision Pro, Microsoft HoloLens, and HTC Vive. While these devices aim to deliver immersive virtual experiences, they introduce a significant challenge in social XR applications by occluding the upper part of the user's face. This occlusion of eyes and eyebrows—features critical for conveying expressions and maintaining eye contact—severely impacts the quality of social interaction in virtual environments.

This limitation particularly affects applications such as virtual reality (VR) teleconferencing \cite{9848469}, collaborative education \cite{rojas2023systematic}, healthcare \cite{ali2023systematic}, and entertainment \cite{simons2023intelligence}, where natural face-to-face communication is essential. Current solutions for addressing HMD occlusion fall into two categories \cite{numan2021generative}: model-based methods that create cartoon-like avatars through additional hardware calibration \cite{lou2019realistic, chen20223d, thies2016facevr}, and image-based methods that aim for photorealistic outputs without extra calibration \cite{numan2021generative, wang2019faithful, zhao2018identity}.

However, existing approaches face significant limitations. State-of-the-art image-based methods are restricted to either single-frame RGB-D inpainting requiring depth cameras \cite{numan2021generative} or basic RGB video inpainting \cite{ghorbani2023expression}. While recent work like the EVI-HRnet model \cite{ghorbani2023expression} effectively removes HMD occlusion using the Learnable Gated Temporal Shift Module (LGTSM) \cite{chang2019learnable}, it lacks the capability to generate complete 3D facial geometry—a crucial requirement for XR telecommunication where users need to view each other from different angles.

Our work addresses these limitations through an efficient approach that creates realistic avatars using only a single RGB camera located in front of the user and HMD eye-tracking data. This significantly reduces hardware complexity while maintaining high-quality results. The key challenge lies in reconstructing 3D face geometry from a single RGB image—a task that requires recovering 3D structures from 2D input without depth information. While neural networks like SynergyNet \cite{wu2021synergy} have shown promise in this domain, integrating such techniques with HMD removal for practical XR applications remains unexplored.

We propose a comprehensive framework that combines RGB video inpainting with 3D face geometry reconstruction, as illustrated in Figure~\ref{fig:teaser}. For development and validation, our research prototype utilizes:

\begin{itemize}
\item Ground truth (GT) RGB frames for training and evaluation,
\item Up to 216 3D landmarks detected from GT frames \cite{wood20223d}, providing detailed facial features including eyebrows, eyes, and eyelid,
\item A single reference frame for preserving identity information.
\end{itemize}

In practical deployment, the system operates with minimum input requirements compared to the current HMD removal approaches:

\begin{itemize}
\item A single external RGB camera capturing the user's face with the HMD,
\item Internal HMD sensors tracking eye and eyebrow movements (providing minimal required landmarks) as the current HMDs mostly support this feature,
\item One reference frame of the user without the HMD (captured during initial setup) for identity preservation.
\end{itemize}

Our framework processes these inputs through several integrated stages. First, a video inpainting step removes the HMD occlusion while preserving facial identity. Then, a dense landmark regression technique refines detected landmarks by integrating 3D Morphable Model (3DMM) semantics into point-based features. Finally, the refined output is processed by SynergyNet to reconstruct 3D facial geometry, producing both inpainted RGB frames and corresponding facial meshes that accurately restore occluded regions.

As we evaluated the model performance with regard to the number of landmarks as input, a key innovation of our approach is its ability to maintain high performance even with sparse landmark data. While our prototype leverages dense landmarks during development, we demonstrate through comprehensive analysis that the framework remains effective with significantly fewer landmarks—a crucial feature for practical deployment where only internal HMD sensors are available for landmark tracking.

The contributions of this work include:
\begin{itemize}
\item A comprehensive framework combining GAN-based RGB video inpainting and face geometry reconstruction to recover occluded facial regions and create 3D face geometry from a single view,
\item An improved facial dense landmark regression technique enhancing the quality of inpainted frames and consequently geometry,
\item A thorough analysis of landmark density impact on reconstruction quality, demonstrating robust performance across varying landmark configurations.
\end{itemize}

The remainder of this manuscript is structured as follows: Section \ref{sec:related_work} reviews related work in HMD removal, video inpainting, and face reconstruction methods. Section \ref{sec:approach} describes the proposed framework in detail. Section \ref{sec:results} presents experimental results both qualitatively and quantitatively, including an ablation study. Section \ref{sec:discussion} discusses the approach in real-world scenarios. Finally, Section \ref{sec:conclusion} concludes with future directions.

\section{Related Work}
\label{sec:related_work}

In this section, we review the related work in HMD removal approaches, which is the application of our proposed framework.
As our approach is based on video inpainting and face reconstruction, we extend our review of the state-of-the-art to also include recent video inpainting methods and face reconstruction from a single view.

\subsection{Head-Mounted Display Removal}
HMDs play a pivotal role in XR environments by enabling immersive experiences. However, their occlusion of the upper face poses significant challenges in social XR applications, e.g. teleconferencing ~\cite{dijkstra2019multi, gunkel2021vrcomm}, electronic learning ~\cite{moutsinas2023application}, and remote collaboration ~\cite{wang2023behere, fidalgo2023magic}, where facial expressions and eye gaze are essential.
Various approaches for HMD removal have been proposed, broadly categorized into model-based and image-based methods.

\subsubsection{Model-based HMD removal}
Model-based approaches typically involve creating person-specific 3D models to reconstruct the occluded facial features, usually with additional hardware inside the HMD requiring calibration for each person. For example, in the so-called \textit{Mask-off} approach~\cite{8797925}, face images in the presence of HMDs have been synthesized using generative models to create a 3D head model. This model-based method reconstructs occluded facial regions by leveraging prerecorded sequences of the face and eye images captured by two near-infrared (IR) cameras inside the HMD.
A recent study introduced a method to reconstruct a fully textured 3D face from a user’s image, followed by training a classifier to assess emotions based on six action units. This approach employed electromyography (EMG) sensors attached to the HMD to track facial muscle movements and expressions, although EMG data was found to lack high accuracy for expression recognition \cite{lou2019realistic}. Another study proposed a person-specific, real-time system capable of reconstructing 3D faces with HMDs and capturing eye gaze. This system uses three IR cameras: two internal cameras for capturing images of the left and right eyes and one external camera for tracking unoccluded facial motion. The external camera, mounted on the HMD, works alongside cell phone sensors to monitor head rotations. Additionally, five IR LED lamps on each side of the device provide consistent illumination for accurate tracking \cite{chen20223d}.

The FaceVR system ~\cite{thies2016facevr} reconstructs the face model when the user is wearing an HMD by placing two IR cameras inside the HMD to capture the eyes, and one RGB-D camera outside the HMD to record the complete face. Similar to this work, a capturing setup is presented in ~\cite{chen2018real} to acquire the facial features in order to create an avatar for the user. This approach also needs IR cameras fixed inside the HMD to properly reconstruct the user’s identity and eye movement, and it requires calibration for each person ~\cite{10.1145/3084363.3085083}.

\subsubsection{Image-based HMD removal}
In contrast, image-based methods aim to produce photorealistic outputs by directly manipulating the image data. These methods focus on retaining the natural appearance and detailed features of the user's face.
Recent works ~\cite{chen2023multi,chen2023dgca,lahiri2020prior,yu2019free} have adopted the concept of generative adversarial networks (GANs) ~\cite{goodfellow2014generative} for image completion, which learns a representative estimate of the distribution of the given training data. Moreover, they also have been employed to solve the HMD removal problem~\cite{gupta2022attention, gupta2022supervision}. In ~\cite{numan2021generative}, the authors introduced a generative RGB-D face completion method for HMD removal, employing GANs to fill in the occluded parts with realistic facial features. This approach focuses on utilizing depth information alongside RGB data to enhance the accuracy of the reconstruction. However, it does not take temporal consistency into consideration and is not directly applicable to video frames. 

In a similar study based on GANs for RGB image inpainting ~\cite{wang2019faithful}, the authors introduced a framework that uses a facial landmark detector. The detected facial landmarks are then passed to a GAN architecture, combined with an occluded RGB image and a reference image. To fill the occluded region in the face image, they first estimate the head pose features from the input image and the reference image, respectively. After that, they provide the input and reference images for a completion network conditioned on the assessed head pose attributes to render an image with the filled missing region. Nevertheless, as the authors stated, their method separately takes each frame for inpainting without considering temporal consistency, so flickering artifacts can be noticed for video applications.

Moreover, in ~\cite{zhao2018identity}, the researchers proposed a GAN-based architecture to preserve the subject’s identity after HMD removal using a reference image of the target subject for face completion. This method employs two discriminators to maintain context consistency and minimize postural errors. As the authors described, jittering can be seen in the transition between frames because the temporal consistency is not explicitly constrained in their formulation, thus leaving it as a future work. 

\subsection{Video inpainting}
Recent advancements in image inpainting have predominantly centered around GANs \cite{yu2018generative,yu2019free}, transformers \cite{wan2021high,liu2021swin}, and diffusion-based models \cite{lugmayr2022repaint,saharia2022palette}. GAN-based models are recognized for generating realistic images \cite{yu2018generative}, while transformer-based models are effective in managing complex tasks due to their capacity to capture contextual information \cite{dosovitskiy2020image}. Diffusion-based models, on the other hand, are distinguished for producing highly detailed textures \cite{ho2020denoising}. However, each approach comes with its own set of challenges: GANs may introduce artifacts in some cases \cite{bau2019seeing}, transformers demand substantial computational resources \cite{dosovitskiy2020image}, and diffusion models are generally slower in processing \cite{ho2020denoising}.

Video inpainting is required for restoring occluded regions in a temporally consistent manner, particularly in applications involving dynamic facial expressions.
Despite the extensive research conducted on image inpainting, video inpainting poses additional challenges that yet need to be fully resolved. Furthermore, the majority of existing studies have focused primarily on object removal and scene inpainting ~\cite{zou2021progressive}, neglecting the specific domain of facial video inpainting involving human subjects, which presents its own set of complexities due to the intricate nature of facial features and the familiarity of faces making a plausible completion more challenging.

Moreover, despite the overall efforts in the field of video inpainting and their promising outcomes, they have mainly revolved around the removal of moving objects or people, i.e. moving masks, across frames, resulting in changing the location of the occluded regions throughout the video sequence. However, in the context of HMD removal, the occluded region remains unchanged in a virtual application, such as teleconferencing, since, when the user wears an HMD, the upper part of the face is continuously occluded. Consequently, the current video inpainting solutions do not address the problem of HMD removal and necessitate adjustments to be practical for XR applications.  
Recently, the EVI-HRnnet method ~\cite{ghorbani2023expression} has been introduced for HMD removal on RGB video frames. 
This method utilized Learnable Gated Temporal Shift Module (LGTSM) ~\cite{chang2019learnable} in their expression-aware video inpainting framework for HMD removal ~\cite{ghorbani2023expression}. This technique enhances the temporal consistency and realism of the inpainted video, which allows for seamless blending of inpainted regions with the rest of the video, preserving the dynamic nature of facial expressions and ensuring a coherent visual experience. However, their work only focused on 2D video inpainting, i.e. the 3D face reconstruction, which is essential for XR telecommunication, was left as future work.

\subsection{Face reconstruction}
Accurate geometry and 3D face reconstruction are essential for realistic face modeling in scenarios where HMDs occlude significant portions of the face. To the best of our knowledge, there are currently no studies that utilize face reconstruction from a single inpainted RGB input for HMD removal in social XR applications. Most existing approaches rely on additional depth cameras to create a person-specific face model of the user.

3D face reconstruction from a single RGB image has been extensively studied in recent years, with notable methods including FFHQ-UV \cite{bai2023ffhq}, 3DDFA-V2 \cite{guo2020towards}, and SynergyNet \cite{wu2021synergy}.

FFHQ-UV is a state-of-the-art UV texture generation method that creates highly detailed and photorealistic textures for 3D face models. By providing high-resolution and realistic textures, FFHQ-UV significantly enhances the visual quality of the final outputs. This technique is useful for achieving a high level of realism in the reconstructed faces, making the models appear lifelike and expressive.

3DDFA-V2 leverages large datasets of annotated facial images and employs convolutional neural networks (CNNs) trained to predict both the 2D facial landmarks and the parameters of a 3D face model simultaneously. This joint prediction enables the system to estimate the 3D geometry of the face directly from the input image. By training on such datasets, these methods aim to capture the underlying relationships between facial appearance in 2D images and corresponding 3D face shapes, enabling accurate and robust reconstruction of facial geometry from single RGB images.

SynergyNet, on the other hand, is a lightweight technique for face geometry reconstruction from a single RGB frame. It captures detailed facial features and landmarks, enabling high-fidelity 3D reconstructions. The integration of SynergyNet as a backbone into our framework ensures that the reconstructed face meshes accurately preserve the user's facial features. This approach builds on the foundation of joint face alignment and 3D face reconstruction.

\section{Approach}
\label{sec:approach}
\begin{figure*}[t]
    \centering
    \includegraphics[width=0.8\linewidth]{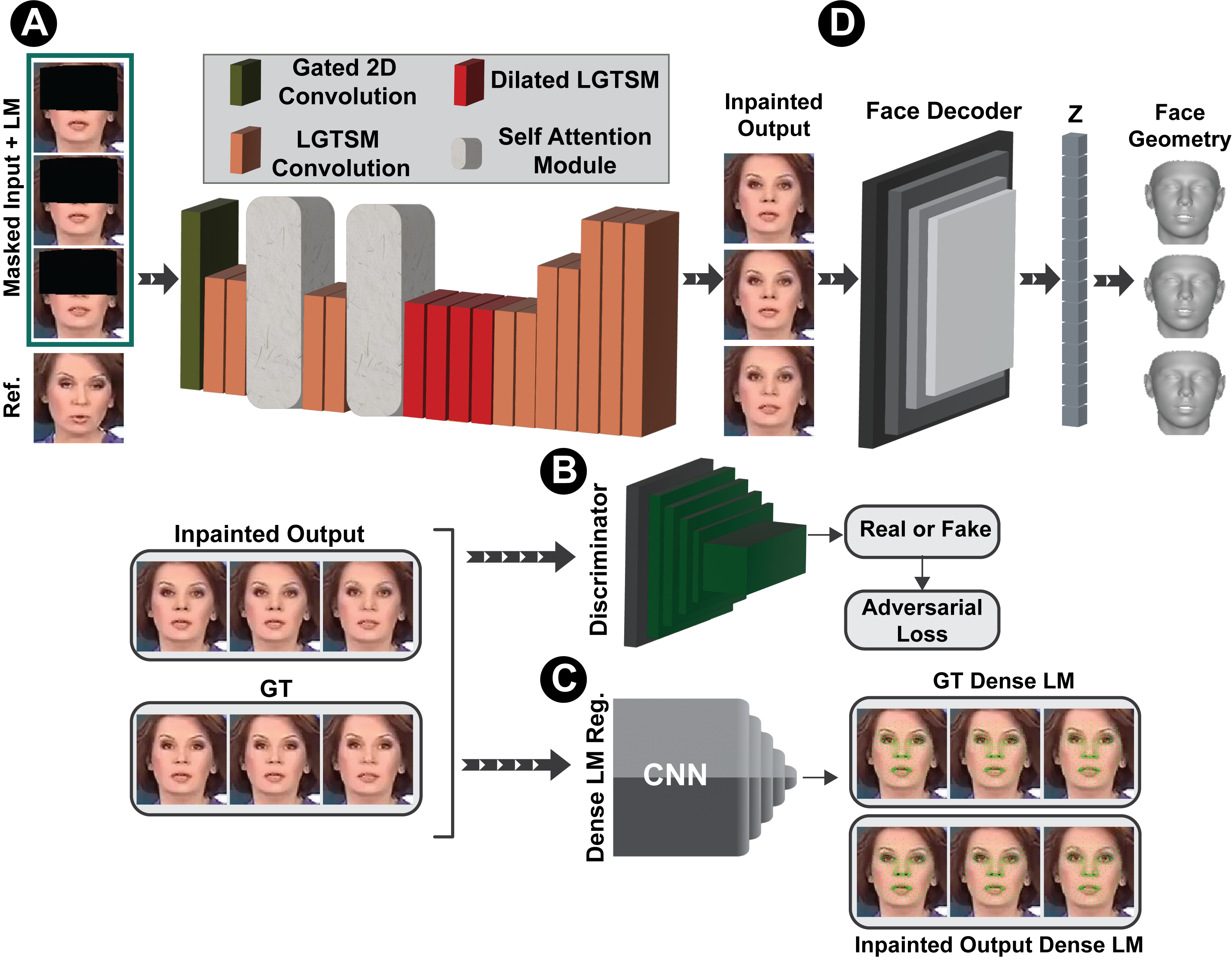}
    \caption{Detailed architecture of the proposed framework. (A) RGB video inpainting using gated 2D convolution, LGTSM convolution, and self-attention modules to fill in occluded facial regions with the help of a single reference image and facial landmarks (LM) detected from GT frames as the model's input. (B) Adversarial learning module with a discriminator network to differentiate real and inpainted frames. (C) Dense landmark regression (Reg.) using a CNN to refine facial landmarks. (D) 3D face geometry reconstruction module to generate face models from inpainted frames based on SynergyNet backbone. (images: TCH (\url{https://www.youtube.com/watch?v=2su-BnPJ9iY})).}
    \label{fig:architecture}
\end{figure*}

\subsection{Framework architecture}

Before detailing our framework, we summarize how it builds upon and differentiates itself from existing state-of-the-art methods. Like prior GAN-based video inpainting methods, we employ adversarial learning and temporal modeling to ensure realism and coherence. However, our method explicitly integrates geometry-aware supervision by combining facial landmark guidance, 3D face reconstruction via SynergyNet, and dense landmark regression (DenseLMLoss), which are absent in most previous approaches. Unlike model-based HMD removal methods, our approach does not require additional hardware such as IR cameras, depth sensors, or EMG devices. Furthermore, compared to 2D-only video inpainting baselines like EVI-HRnet, our framework jointly restores both 2D facial appearance and full 3D facial geometry. Finally, we address practical XR constraints by demonstrating robustness even under sparse landmark availability, an important limitation not covered by previous works.

Our framework for removing HMD occlusions and reconstructing 3D face geometry in XR applications consists of four sequential stages, as illustrated in Figure~\ref{fig:architecture}.

The process begins with the preparation of input data, which includes ground-truth (GT) RGB frames, masked frames simulating HMD occlusion, and a single unoccluded reference frame. 
Dense facial landmarks are detected from the GT frames to provide initial geometric guidance. 
These landmarks, along with the masked frames and the reference frame, are then fed into our RGB video inpainting model, EVI-HRnet~\cite{ghorbani2023expression}.

\textbf{Step A: RGB Video Inpainting.}  
EVI-HRnet performs the inpainting task by filling the occluded facial areas, leveraging the structural cues from dense landmarks and visual details from the reference frame. 
This step ensures the preservation of facial identity and expression continuity.

\textbf{Step B: Adversarial Supervision.}  
An adversarial discriminator network is used during training to differentiate between real and inpainted frames.
The adversarial loss encourages the generator to produce photorealistic and temporally coherent sequences.

\textbf{Step C: Dense Landmark Regression.}  
After inpainting, a CNN-based dense landmark regression module refines the facial landmarks on the inpainted frames.
This module incorporates 3D Morphable Model (3DMM) semantics into point-based features to enforce geometric consistency.

\textbf{Step D: 3D Face Geometry Reconstruction.}  
Finally, the refined landmarks and the inpainted frames are passed to the SynergyNet~\cite{wu2021synergy} backbone.
SynergyNet regresses full 3DMM parameters and reconstructs detailed 3D face meshes with realistic shape and expression information.

It is important to note that while the dense landmarks extracted from GT frames guide the inpainting process, the final 3D face reconstruction is performed solely on the inpainted outputs, ensuring consistency and realism without relying on original ground-truth frames.

The following subsections provide a comprehensive description of each stage of the framework, following the sequence in Figure~\ref{fig:architecture}: input data preparation, RGB video inpainting, geometry reconstruction, and optimization techniques.

\subsection{Input data}

The inputs to our pipeline consist of binary masks simulating the HMD, a single reference frame without masks, and GT frames. We apply the HMD masks to GT frames to create the occluded frames. From the reference frame, we only use the missing information in the corresponding occluded area of masked frames. 
Dense facial landmarks are detected from the GT frames using the MediaPipe Face Landmarker ~\cite{wood20223d}. The MediaPipe Face Landmarker outputs 3D face landmarks by estimating normalized 2D coordinates (x, y) along with a relative depth (z) value for each landmark point. Although this provides 3D information, it does not constitute a full 3D mesh reconstruction. Instead, it generates 478 individual 3D landmark points based on learned correlations between 2D facial features and typical 3D face structure. The MediaPipe Face Landmarker uses a series of models for this purpose. The first model detects faces, the second locates landmarks on the detected faces, and the third identifies facial features and expressions. The face mesh model outputs an estimate of 478 3D face landmarks, providing a comprehensive mapping of the face. It estimates the landmark positions on the 2D image plane and provides an additional relative depth value for each landmark. This depth estimation is based on learned correlations between 2D facial features and their typical 3D structure.

The complete 3D face mesh used in our pipeline is later reconstructed by SynergyNet, which generates dense vertex coordinates (x, y, z) from the inpainted RGB frames.

These detected landmarks in the masked area (up to 216 including eyebrows and eyes), along with the masked frames and the reference frame, are then fed into our model. The inclusion of an RGB reference face frame into the model’s input is crucial for addressing the HMD removal problem while ensuring the preservation of the person's identity in the final face model. 
To establish the reference frame, we select the first frame in the sequence as the single reference image. The reference frame can also be selected randomly from the available frames, providing flexibility while maintaining the necessary input for the generator to perform its inpainting task accurately.
To focus on the occluded part (HMD occlusion) during inpainting, we zero out the pixels in the reference frame that are outside the HMD mask area. This ensures that the model concentrates on reconstructing the occluded facial regions, leveraging the dense landmarks to achieve precise and realistic results.

\subsection{RGB video inpainting}

The RGB inpainting of our framework is centered around the EVI-HRnet ~\cite{ghorbani2023expression}. This architecture includes an attention-based LGTSM generator and a temporal shift module (TSM) patch discriminator, which are trained via an adversarial process, as shown in Figure~\ref{fig:architecture}. In this process, the goal of the generator is to correctly inpaint a masked frame using a single unoccluded reference image and facial landmarks. The aim of the discriminator is to specify whether the final inpainted frame is real or fake.
The LGTSM layers ~\cite{chang2019learnable} in EVI-HRnet optimize 2D convolutions by intelligently shifting input channels to their temporal neighbors thus enhancing temporal understanding crucial for video inpainting tasks. This design choice eliminates the need for additional parameters from 3D convolutions or optical flow data, resulting in a lightweight yet high-performance architecture.
This model is integrated with an attention mechanism \cite{zhang2019self}, which empowers the network to focus on diverse parts of the input data with regard to occlusion movement, particularly improving global context understanding and capturing non-local features within the feature maps. 

More in detail, the generator in the inpainting model consists of 13 convolution layers with the gated TSM, and uses attention-based down-sampling, dilation, and up-sampling as depicted in Figure~\ref{fig:architecture} (A). Self-attention layers are inserted after the LGTSM blocks in the downsampling path of the decoder. They compute attention weights, enabling the network to capture long-range spatial dependencies and temporal relationships across video frames, which are essential for improving temporal consistency in the inpainted results. Self-attention layers are strategically positioned to compute attention weights, allowing the network to capture spatial relationships, dependencies, and feature information within the input feature maps. 

In the adversarial learning process, the discriminator evaluates inpainted frames against GT frames, compelling the generator to accurately fill occluded areas. This involves six 2D convolution layers with TSM, ensuring a comprehensive evaluation of the generated frames. The discriminator further ensures the consistency of highly detailed features across distant portions of the frame, enhancing the overall visual fidelity.

\subsection{Geometry reconstruction}

We propose to use SynergyNet \cite{wu2021synergy} which serves as a lightweight backbone in our proposed framework to reconstruct the 3D face geometry from inpainted RGB frames acquired from EVI-HRnet. By incorporating SynergyNet, we can effectively bridge the gap between RGB inpainted frames and 3D face models, ensuring that the reconstructed geometries are both accurate and detailed.

Importantly, the depth information (i.e., the z-coordinate) of the 3D face model is not estimated through a separate depth network. Instead, it is directly derived from the 3D Morphable Model (3DMM) parameters regressed by SynergyNet. These parameters encode the full 3D structure of the face, including shape, expression, and pose components, allowing the generation of a 3D face mesh with per-vertex $(x, y, z)$ coordinates. Thus, depth is inherently incorporated into the reconstructed mesh geometry.

The architecture of SynergyNet is structured into two main stages, as shown in Figure~\ref{fig:architecture}. In the first stage, the network tries to regress 3D Morphable Model (3DMM) parameters from images, which are then used to construct 3D face meshes. This stage also involves the extraction of landmarks by querying associated indices. 

Following this, a landmark refinement module is introduced to aggregate 3DMM semantics and integrate them into point-based features, producing refined 3D landmarks.

The second stage of SynergyNet involves a landmark-to-3DMM module, which predicts 3DMM parameters from the refined 3D landmarks. Specifically, the predicted 1D parameters refer to the 3D Morphable Model (3DMM) coefficients, including identity, expression, and pose vectors, which are then used to reconstruct the full 3D face mesh with dense (x, y, z) vertex coordinates. This feedback step further refines the embedded facial geometry, ensuring high fidelity both in surface topology and depth structure, without requiring an explicit depth regression branch.

This represents a reverse direction in representation compared to the first stage, moving from 3D landmarks back to 1D parameters. This step is leveraged to regress embedded facial geometry based on the sparse landmarks, ensuring that the final 3D face model is highly detailed and accurate. By integrating these steps, SynergyNet effectively captures and refines facial geometry, making it a robust backbone for 3D face reconstruction in our framework.

It is noteworthy that, due to the unavailability of the person's face mesh model in real-world scenarios, the GT face geometry models are also generated from the original RGB video frames without HMD occlusion, using the SynergyNet method. This helps the model to learn the desired features during optimization and training of the whole framework.

\subsection{Loss functions}

Our model employs a combination of diverse loss functions for effective convergence. The L1 Reconstruction Loss (ReconLoss) emphasizes pixel-wise accuracy, measuring the fidelity of inpainted frames concerning GT frames. The VGG Loss (VGGLoss) based on ImageNet captures perceptual differences by utilizing a pre-trained VGG network on ImageNet, providing insights into high-level features ~\cite{simonyan2014very,russakovsky2015imagenet}. The Style Loss (StyleLoss) ~\cite{gatys2015neural} ensures the preservation of stylistic features in the inpainted frames. The Wasserstein GAN Adversarial Loss (AdvLoss) further guides the generator to create realistic inpainted frames by misleading the discriminator ~\cite{arjovsky2017wasserstein}. Moreover, the facial expression recognition Loss (FERLoss) evaluates the model's performance in recognizing multiple facial expression classes designed based on ~\cite{savchenko2022video}, ensuring an accurate depiction of emotions in the inpainted frames. More in detail, for FER loss, we calculate the FER score for eight facial expression classes (namely surprise, angry, sad, contempt, disgust, fear, neutral, and happy) using an EfficientNet model ~\cite{savchenko2022video} trained on the AFEW dataset ~\cite{dhall2012collecting}. 

Additionally, we propose a dense landmark regression method (DenseLMLoss) that ensures accurate reconstruction of the final face geometry. In our method, we utilize the \textbf{Huber loss} function to evaluate the fidelity of landmark predictions on the inpainted results compared to the GT. The Huber loss is a robust loss function that combines the advantages of L1 and L2 norms. It reduces the influence of outliers by behaving quadratically for small errors and linearly for large errors, maintaining a stable gradient while avoiding large deviations that might occur in the presence of outliers.

The Huber loss for each landmark is calculated as follows:

\begin{equation}
L_{\text{Huber}}(a, b) = 
\begin{cases} 
\frac{1}{2}(a - b)^2 & \text{for } |a - b| \leq \delta \\
\delta |a - b| - \frac{1}{2} \delta^2 & \text{otherwise}
\end{cases}
\end{equation}

Where \( a \) and \( b \) represent the predicted and GT landmark positions, respectively, and \( \delta \) is a threshold that determines the point at which the loss function transitions from quadratic to linear.

In the context of dense landmark regression, this loss is computed by comparing the predicted landmark positions on the inpainted frames with the corresponding landmarks in the GT frames. The loss function is applied to all landmarks, and the average loss across the batch is computed. Specifically, the \textbf{DenseLMLoss} method involves the following steps:

\begin{enumerate}
    \item \textbf{Landmark Extraction}: The landmarks are first extracted from both the GT and the model's inpainted frames using the MediaPipe Face Landmarker ~\cite{wood20223d}, which outputs a detailed map of 478 face landmarks. The landmarks are expected to be in 3D space with coordinates \( (x, y, z) \), where \( x \) and \( y \) represent normalized image coordinates, and \( z \) corresponds to the depth of the landmark.
    
    \item \textbf{Huber Loss Calculation}: The Huber loss is computed for each landmark across the entire batch, frame by frame, and landmark by landmark. The calculated loss is then averaged over all landmarks, yielding the final DenseLMLoss.
    
    \item \textbf{Final Loss}: The final loss is given by:
    
\begin{equation}
L_{\text{DenseLM}} = \frac{1}{N} \sum_{i=1}^{N} L_{\text{Huber}}(\hat{p}_i, p_i)
\end{equation}

Where \( \hat{p}_i \) is the predicted landmark position, \( p_i \) is the corresponding GT position, and \( N \) is the total number of landmarks in the batch. This loss is then used to optimize the model, ensuring that the landmarks are accurately predicted, leading to a more precise face geometry reconstruction.
\end{enumerate}

The input-output pairs used in each loss function are defined as follows:
\begin{itemize}
    \item ReconLoss: between inpainted frames and GT frames.
    \item VGGLoss and StyleLoss: between VGG feature representations of inpainted frames and GT frames.
    \item AdvLoss: discriminator distinguishes between GT RGB frames and generated inpainted frames.
    \item FERLoss: predicted facial expression classes from inpainted frames vs. GT emotion labels.
    \item DenseLMLoss: predicted 3D landmarks extracted from SynergyNet output (inpainted frames) vs. GT 3D landmarks extracted by MediaPipe from the original non-occluded frames.
\end{itemize}

To further refine the training process, each loss computation is assigned a scaling factor, emphasizing its relative importance in the overall loss function. The overall loss function for training the model can be expressed as:

\begin{equation}
\begin{split}
    \text{Total Loss} = & \lambda_{\text{AdvLoss}} \cdot \text{AdvLoss} + \lambda_{\text{StyleLoss}} \cdot \text{StyleLoss} \\
    & + \lambda_{\text{VGGLoss}} \cdot \text{VGGLoss} + \lambda_{\text{FERLoss}} \cdot \text{FERLoss} \\
    & + \lambda_{\text{ReconLoss}} \cdot \text{ReconLoss} + \lambda_{\text{DenseLMLoss}} \cdot \text{DenseLMLoss}
\end{split}
\label{equ:loss}
\end{equation}

where $\lambda_{\text{AdvLoss}}$, $\lambda_{\text{FERLoss}}$, $\lambda_{\text{StyleLoss}}$, $\lambda_{\text{VGGLoss}}$, $\lambda_{\text{ReconLoss}}$, and $\lambda_{\text{DenseLMLoss}}$ are the weights for AdvLoss, FERLoss, StyleLoss, VGGLoss, ReconLoss, and DenseLMLoss, respectively.

Each loss compares clearly defined outputs and ground truths, improving both inpainting realism and geometric fidelity.

\paragraph{Training Protocol:} The training of our model follows a two-stage strategy for stability and optimal performance:
\begin{itemize}
    \item \textbf{Stage 1: Generator pretraining.} We first pretrain the generator network for 100 epochs using all losses except the adversarial loss (i.e., $\lambda_{\text{AdvLoss}} = 0$). This warm-up phase helps the generator to learn basic inpainting and facial feature reconstruction before adversarial learning is introduced.
    \item \textbf{Stage 2: Joint training.} After generator pretraining, the discriminator is activated, and the full adversarial training begins. Both the generator and discriminator are jointly optimized according to the complete loss formulation in Equation~\ref{equ:loss}, with all loss weights $\lambda$ kept fixed throughout.
\end{itemize}
\textbf{Pre-trained components:} The VGG feature extractor is initialized with weights pre-trained on ImageNet, and the FER classifier is based on an EfficientNet trained on the AFEW dataset~\cite{dhall2012collecting}. 
\textbf{SynergyNet fine-tuning.} During Stage 1, the SynergyNet module is frozen to prevent destabilization from noisy inpainting outputs. In Stage 2, it is fine-tuned together with the generator to adapt its latent 3D representation to the evolving inpainted frames and refined dense landmark maps.

\section{Experimental Results}
\label{sec:results}

In this section, we present our experimental results for both RGB video inpainting and 3D geometry reconstruction. For the RGB video inpainting component, we compare against EVI-HRnet~\cite{ghorbani2023expression}, currently the only available video inpainting model specifically designed for HMD removal. Due to the lack of publicly available implementations of other HMD removal methods that handle temporal features, we extend our comparison to include general facial video inpainting methods: LGTSM \cite{chang2019learnable} and CombCN \cite{wang2019video}. CombCN employs a two-stage approach combining 3D fully convolutional architecture for temporal structure inference with 2D convolutions for spatial detail recovery.

The reference frame used for inpainting is excluded from the masked inputs in our method and is incorporated into the model’s architecture as explicit identity guidance. For baseline methods, the first frame is left unmasked during training to ensure a fair input setup, but no special mechanism for leveraging it as a reference is employed.

For 3D face geometry reconstruction, direct comparisons with existing HMD removal approaches are not feasible as they typically require person-specific calibration and additional hardware (e.g., internal cameras, depth sensors). Therefore, we conduct a comprehensive ablation study of our framework using different numbers of facial landmarks to evaluate the relationship between landmark density and reconstruction quality.

This section first describes implementation details and framework setup, followed by dataset description and preparation. We then present the quantitative evaluation of RGB video inpainting compared to baseline methods, followed by the analysis of 3D geometry reconstruction through ablation studies on landmark density. Finally, we provide a qualitative assessment of both inpainted frames and reconstructed 3D face models.

\begin{table}[ht]
\caption{Quantitative evaluation results of EVI-HRnet, CombCN, LGTSM, and our model with varying numbers of landmarks (LM). The metrics are averaged results of the FaceForensics \cite{rossler2018faceforensics} test set. The best results are in {\bf bold}, and the second best in {\it italic}.} 
\label{tab:metrics}
\begin{center}       
\begin{tabular}{|l|c|c|c|c|c|} 
\hline
\rule[-1ex]{0pt}{3.5ex}  \textbf{Model} & \textbf{FID\(\downarrow\)} & \textbf{MSE\(\downarrow\)} & \textbf{LPIPS\(\downarrow\)} & \textbf{SSIM\(\uparrow\)} & \textbf{PSNR\(\uparrow\)}  \\
\hline\hline
\rule[-1ex]{0pt}{3.5ex}  EVI-HRnet (68 LM) & 0.7675 & 0.0030 & 0.0508 & 0.8986 & 25.83  \\
\hline
\rule[-1ex]{0pt}{3.5ex}  CombCN & 0.8721 & 0.0035 & 0.0590 & 0.8812 & 24.67  \\
\hline
\rule[-1ex]{0pt}{3.5ex}  LGTSM & 0.9124 & 0.0037 & 0.0615 & 0.8754 & 24.12  \\
\hline
\rule[-1ex]{0pt}{3.5ex}  Ours (216 LM) & \textbf{0.6785} & \textbf{0.00204} & \textbf{0.0442} & \textbf{0.9212} & \textbf{27.57}  \\
\hline
\rule[-1ex]{0pt}{3.5ex}  Ours (68 LM) & {\it 0.7076} & {\it 0.00230} & {\it 0.0501} & {\it 0.9150} & {\it 27.16}  \\
\hline
\rule[-1ex]{0pt}{3.5ex}  Ours (20 LM) & {0.7438} & {0.00231} & {0.0504} & {0.9112} & {27.15}  \\
\hline
\rule[-1ex]{0pt}{3.5ex}  Ours (10 LM) & {0.8016} & {0.00232} & {0.0575} & {0.9098} & {27.05}  \\
\hline
\end{tabular}
\end{center}
\end{table}

\subsection{Implementation details and setup}
We utilize PyTorch 1.10.0 for the implementation of the network. We employ a kernel size of $5\times5$ for the first convolution layer, a kernel size of $4\times4$ with stride 2 for the down-sampling layers, a kernel size of $3\times3$ with dilation of 2, 4, 8, 16 for the dilated layers, and a kernel size of $3\times3$  for other convolution layers, similar to ~\cite{chang2019learnable}. For the attention layers, we utilize a kernel size of $1\times1$. As an activation function, we use the LeakyReLU. The Adam optimizer with a learning rate of $9.6 \times 10^{-5}$ is employed for training. 

Finally, we set the weights of the overall loss function in \ref{equ:loss} as follows:

\begin{equation}
\lambda_{\text{AdvLoss}} = 1, \quad \lambda_{\text{FERLoss}} = 2, \quad \lambda_{\text{StyleLoss}} = 10, \quad \lambda_{\text{VGGLoss}} = 1, \quad \lambda_{\text{ReconLoss}} = 1, \quad \lambda_{\text{DenseLMLoss}} = 1
\end{equation}

for AdvLoss, FERLoss, StyleLoss, VGGLoss, ReconLoss, and DenseLMLoss, respectively.

As clarified in last section, the ground-truth (GT) 3D face geometry models used for evaluation are generated by applying SynergyNet to the original, unoccluded RGB frames. These GT meshes are utilized solely for visualization and comparison of reconstruction results. 
Despite the larger model size, the added modules enable enhanced capabilities, such as identity preservation across frames and detailed 3D mesh reconstruction, which are not available in baseline methods. Furthermore, the training procedure is modular: the generator is first pretrained independently, adversarial learning is introduced afterward, and fine-tuning of the geometry branch (SynergyNet) is performed only after stable inpainting performance is achieved.

\subsection{Video dataset}
We utilize the FaceForensics \cite{rossler2018faceforensics} dataset, which comprises 1,004 videos, including more than 500,000 frames featuring the faces of newscasters collected from YouTube. Most of these videos contain frontal faces cropped to a size of $128\times128$ pixels, making them ideal for training learning-based models. We use 150 videos for testing, each with a duration of 32 frames, while the remaining videos are allocated for training the models.

While FaceForensics offers a relatively large number of identities and high-quality video sequences, we acknowledge that the dataset predominantly features individuals recorded in professional, well-lit settings (e.g., news presenters and interviewees). As such, it provides limited diversity in terms of age, ethnicity, lighting variations, and casual expressions. Although it remains the largest publicly available benchmark suited for controlled HMD occlusion experiments, future work will focus on expanding training and evaluation to datasets that better capture broader demographic variation and more unconstrained environments.

To simulate HMDs on the faces in the video frames, we first created binary masks of an HMD. These masks were then applied to all GT frames, serving as input data for the networks. The HMD masks used in this study are provided on the project page.

\begin{figure}[h]
\centering
\includegraphics[width=\textwidth]{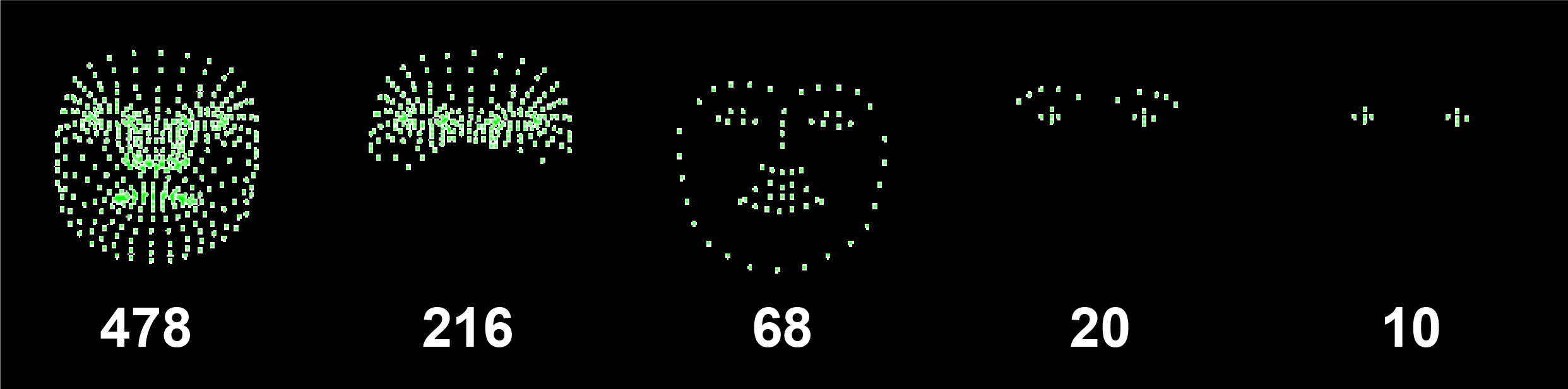}
\caption{Progressive reduction in facial landmark density. From left to right: Original MediaPipe 478-point dense 3D landmarks shown for reference, experimental configurations using 216-point set within the masked region, standard 68-point 2D landmarks (mapping jaw, eyebrows, eyes, nose, and mouth), focused 20-point configuration (eyes and eyebrows), and minimal 10-point set (eyelids only). This systematic reduction enables analysis of landmark density impact on inpainting performance.}
\label{fig:landmarks}
\end{figure}

\subsection{Quantitative results}
\subsubsection{RGB video inpainting}

We present the quantitative evaluation results of our proposed framework, which integrates RGB video inpainting and 3D face geometry reconstruction, comparing it against the EVI-HRnet model as well as the LGTSM and CombCN models. We use mean square error (MSE), peak-signal-to-noise ratio (PSNR),  and structural similarity index (SSIM) ~\cite{wang2004image} to evaluate the image quality in all models. Additionally, we also report Learned Perceptual Image Patch Similarity (LPIPS) ~\cite{zhang2018unreasonable}, and Fréchet inception distance (FID) ~\cite{heusel2017gans} score as our evaluation metrics, which have been demonstrated to align remarkably well with human judgments of image similarities.

While LPIPS and FID are primarily spatial metrics, they have been widely adopted for video tasks due to their implicit sensitivity to temporal artifacts across frame sequences. To further assist the reader in evaluating temporal consistency, we will release output RGB video clips and meshes on our public GitHub page.

Our experimental framework evaluates the impact of facial landmark density on model performance through a systematic comparison of different configurations. While the original MediaPipe model provides 478 dense 3D facial landmarks, we strategically reduce the number of landmarks to minimize computational overhead and focus on features most relevant to HMD occlusion. Our experiments examine four configurations: 216 landmarks within the masked facial region, standard 68-point landmarks (17 points for jaw outline, 10 points for eyebrows, 12 points for eyes, 9 points for nose, and 20 points for mouth), focused 20-point set (5 points per eyelid, 5 points per eyebrow), and minimal 10-point configuration tracking eyelid contours (5 points per eye). The quantitative results of RGB video inpainting are presented in Table~\ref{tab:metrics}. Figure~\ref{fig:landmarks} demonstrates this progressive reduction in landmark density.

The proposed pipeline demonstrates significant improvements over the EVI-HRnet model. For the same number of landmarks (68), our model achieves a lower FID score (0.7076 vs. 0.7675), MSE (0.00230 vs. 0.0030), and LPIPS (0.0501 vs. 0.0508), along with higher SSIM (0.9150 vs. 0.8986) and PSNR (27.16 vs. 25.83). These enhancements highlight the efficacy of dense landmark regression in improving the quality and realism of inpainted frames and reconstructed 3D face geometry.

The proposed pipeline with 216 dense landmarks yields the best results across all metrics, demonstrating the benefits of using a higher number of landmarks for detailed face geometry reconstruction. Specifically, it achieves the lowest FID (0.6785), MSE (0.00204), and LPIPS (0.0442), and the highest SSIM (0.9212) and PSNR (27.57). In contrast, reducing the number of landmarks to 20 and 10, while still showing improvements over the EVI-HRnet, results in slightly lower performance. Notably, the model with 10 landmarks shows lower performance in FID and LPIPS, which are metrics closely related to human visual judgment, despite having better SSIM and PSNR scores. This indicates that a minimum number of landmarks is necessary to achieve an acceptable level of performance.

Our proposed pipeline outperforms both CombCN and LGTSM across all evaluation metrics. The CombCN model achieves an FID of 0.8721, MSE of 0.0035, LPIPS of 0.0590, SSIM of 0.8812, and PSNR of 24.67. Similarly, the LGTSM model has an FID of 0.9124, MSE of 0.0037, LPIPS of 0.0615, SSIM of 0.8754, and PSNR of 24.12. These results indicate that the proposed new pipeline with dense landmark regression, especially with 216 landmarks, provides superior performance, generating more realistic and consistent video frames.

In summary, the proposed approach significantly improves upon the EVI-HRnet model and outperforms the CombCN and LGTSM models. The incorporation of dense landmark regression and the use of 216 landmarks are particularly beneficial, leading to the best performance in terms of FID, MSE, LPIPS, SSIM, and PSNR. These findings underscore the importance of detailed landmarks and the enhanced capability of our framework to produce realistic, consistent, and high-quality 3D face reconstructions in XR applications. It is worth mentioning that the lower performance with 10 landmarks indicates that, even with our pipeline and its components, a minimum number of landmarks is required to maintain high performance across all metrics.

\subsubsection{Face geometry reconstruction}

We evaluated the 3D face reconstructions produced by our method with various landmark configurations: 216 landmarks, 68 landmarks, 20 landmarks, and 10 landmarks. As shown in our RGB quality evaluation, the proposed pipeline with 216 dense landmarks yields the best results across all evaluation metrics, demonstrating the advantages of using a larger number of landmarks to achieve more accurate face geometry reconstruction. 

The impact of landmark density on the accuracy of 3D geometry is quantitatively assessed using three metrics: Average Chamfer Distance, Average RMS Error, and Average Mean Hausdorff Distance. 

The average Chamfer Distance outlines the precision with which our method reconstructs the surface geometry of 3D models compared to GT 3D models.

The average RMS Error, by quantifying the mean discrepancy between the predicted and actual vertex positions, underscores the precision of our model in rendering the facial features accurately, ensuring the reconstructed models faithfully represent the original subjects.

Average Mean Hausdorff Distance, on the other hand, provides insight into the maximal extent of deviation between the reconstructed models and their real counterparts (GT 3D models). The results are presented in Table~\ref{tab:comparison}, where we observe a clear trend: as the number of landmarks increases, the accuracy of the reconstruction improves, resulting in lower error values across all metrics.

\begin{table}[ht]
\caption{Quantitative comparison of 3D reconstruction results of our model with GT geometry for different landmark configurations. All distance metrics are measured in normalized pixel units relative to the input frame resolution. Lower values indicate better reconstruction quality.}
\label{tab:comparison}
\begin{center}
\begin{tabular}{|l|c|c|c|}
\hline
\rule[-1ex]{0pt}{3.5ex} \textbf{Method} & \textbf{Average Chamfer Distance} & \textbf{Average RMS Error} & \textbf{Average Hausdorff Distance} \\
\hline\hline
\rule[-1ex]{0pt}{3.5ex}  216 LM & 1.721 & 1.121 & 0.984 \\
\hline
\rule[-1ex]{0pt}{3.5ex}  68 LM & 2.015 & 1.243 & 1.148 \\
\hline
\rule[-1ex]{0pt}{3.5ex}  20 LM & 2.512 & 1.399 & 1.243 \\
\hline
\rule[-1ex]{0pt}{3.5ex}  10 LM & 2.754 & 1.487 & 1.452 \\
\hline
\end{tabular}
\end{center}
\end{table}

From the results in Table~\ref{tab:comparison}, we observe that the model with 216 landmarks achieves the best performance with the lowest values in all three metrics: Average Chamfer Distance (1.721), Average RMS Error (1.121), and Average Hausdorff Distance (0.984). This reflects a higher level of precision in reconstructing the 3D facial geometry, aligning closely with the ground truth (GT) models generated by SynergyNet.

In comparison, when the number of landmarks is reduced to 20 or 10, there is a noticeable degradation in performance. Specifically, the model with 20 landmarks produces higher error values: Chamfer Distance (2.512), RMS Error (1.399), and Hausdorff Distance (1.243). This suggests that with fewer landmarks, the model struggles to capture finer facial details, leading to a less accurate reconstruction. The model with only 10 landmarks further exacerbates this trend, with even higher error metrics: Chamfer Distance (2.934), RMS Error (1.487), and Hausdorff Distance (1.452). 
However, it is noteworthy that the degradation is gradual, indicating that the model remains robust even with very sparse landmark inputs. The increase in error across different metrics is relatively smooth, demonstrating the method’s robustness to varying levels of input sparsity.

These results align with our RGB quality observations. As shown by the RGB-based metrics (FID, MSE, LPIPS, SSIM, and PSNR), the model with 216 landmarks outperforms others.

In conclusion, these results emphasize the importance of using a sufficient number of landmarks to achieve both high-quality RGB results and precise 3D geometry reconstruction. The model's performance improves as the landmark count increases, reducing error metrics such as Chamfer Distance, RMS Error, and Hausdorff Distance, and leading to more realistic and accurate 3D reconstructions of the face.

\begin{figure}[h!]
  \centering
  \begin{minipage}{0.48\textwidth}
    \centering
    \includegraphics[scale=0.4]{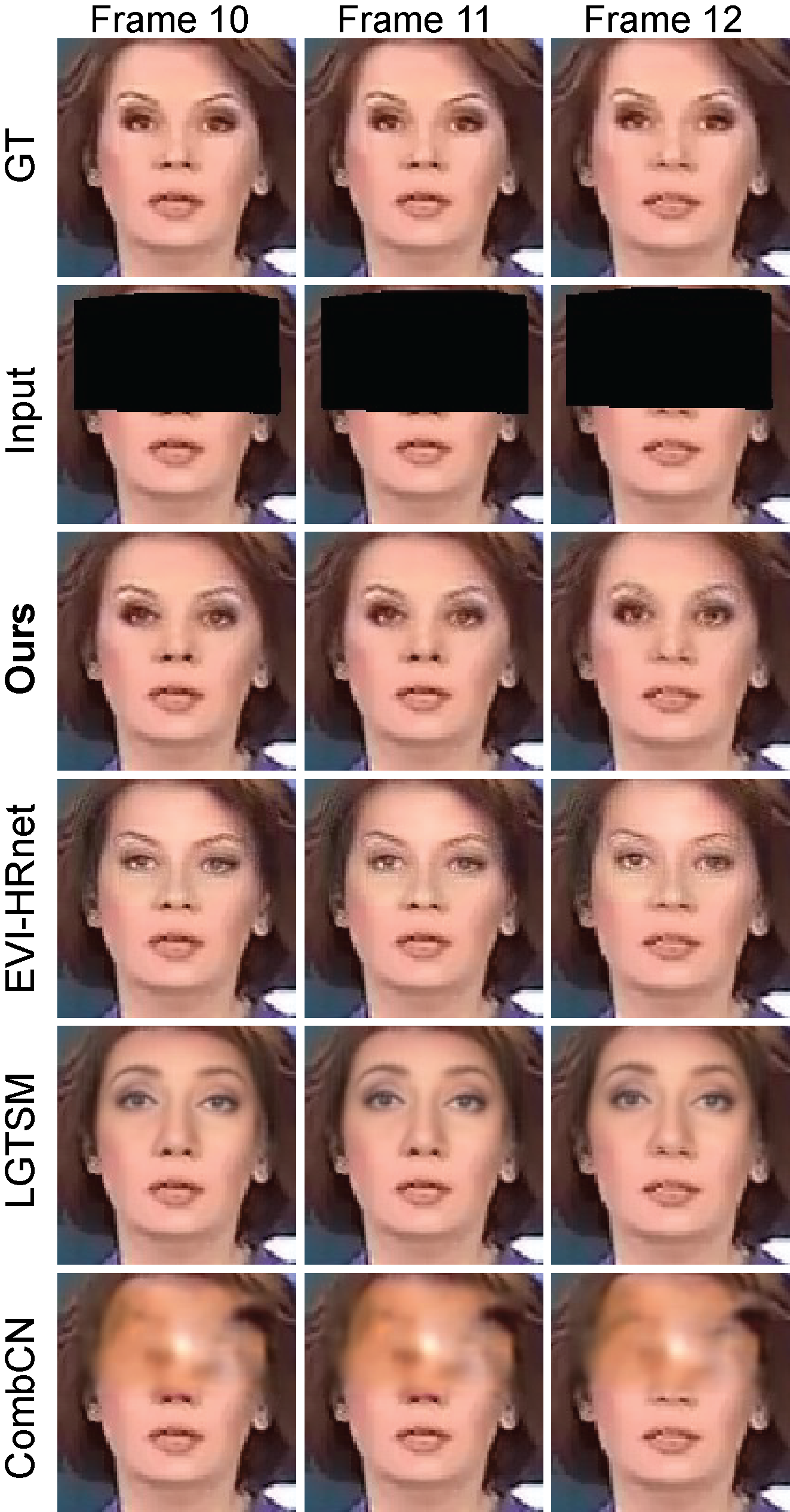}
    \caption{Sample of the qualitative results of FaceForensics test set with HMD masks, and their GT and inputs. The inpainted frames are selected from the results of our model, EVI-HRnet, LGTSM, and CombCN. (images: TCH (\url{https://www.youtube.com/watch?v=2su-BnPJ9iY})).}
    \label{qual_res}
  \end{minipage}%
  \hfill
  \begin{minipage}{0.48\textwidth}
    \centering
    \includegraphics[scale=0.35]{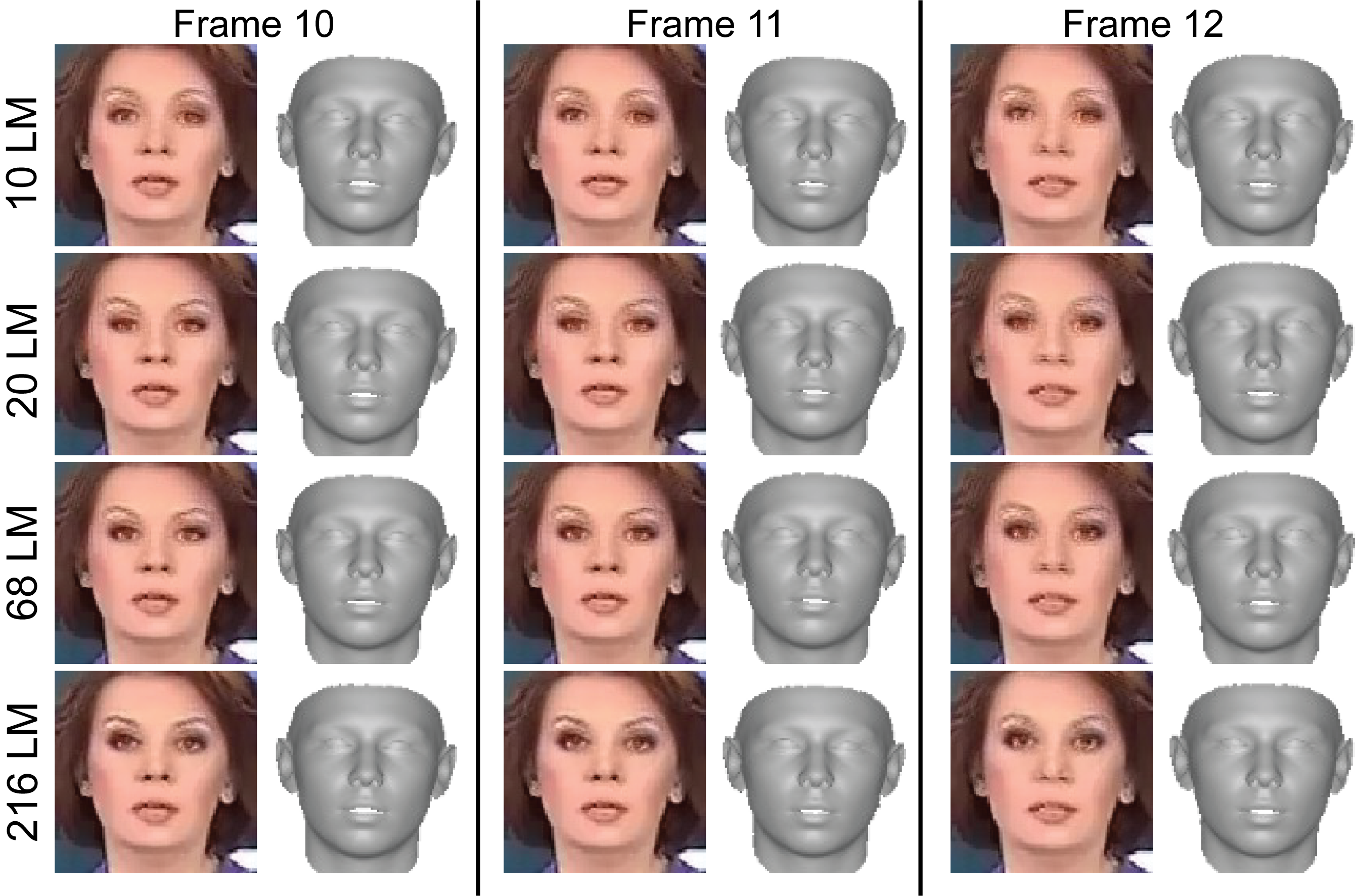}
    \caption{A sample of qualitative ablation comparison of our proposed method with regards to the number of landmarks (LM) on the FaceForensics \cite{rossler2018faceforensics} test set. The final face geometry is also depicted. (images: TCH (\url{https://www.youtube.com/watch?v=2su-BnPJ9iY})).}
    \label{Ablation}
  \end{minipage}
\end{figure}

\subsection{Qualitative results}
\subsubsection{RGB video inpainting}
To evaluate the effectiveness of our proposed geometry-aware inpainting method, we provide qualitative comparisons against existing methods, including EVI-HRnet, CombCN, and LGTSM, as shown in Figure~\ref{qual_res}. These comparisons highlight the capability of each model to reconstruct occluded facial regions and recover fine details, which are essential for producing visually realistic outputs in the context of HMD removal for social XR.

Recovering fine details, such as subtle changes in facial expressions, eyebrow movements, and realistic eyelash rendering, is critical for creating a believable experience in XR environments. Our proposed model demonstrates superior performance compared to all other methods by using a geometry-aware inpainting process. This process effectively addresses HMD occlusion while preserving the user's facial identity and expressions, leading to highly realistic inpainted frames, which are aligned with our quantitative analysis. The ability of our model to recover accurate eye details, maintain facial features, and ensure smooth transitions between expressions contributes significantly to the overall realism and authenticity of the outputs, making the inpainted frames more visually compelling and believable, as shown in Figure~\ref{qual_res}.

In contrast, the EVI-HRnet model, although it utilizes facial landmarks and a single reference image to enhance inpainting, does not achieve the same level of detail recovery as our proposed model. While EVI-HRnet can produce visually realistic outputs to some extent, it falls short in further capturing face geometry and the finer details that our model handles more effectively.

The CombCN model exhibits noticeable limitations, particularly in restoring the eye regions, often failing to reconstruct the eyes fully and accurately. Meanwhile, the LGTSM model shows an improvement over CombCN by successfully recovering the eyes in every frame. However, it suffers from a consistent issue where the eyes are mispositioned, leading to inpainted frames that fail to reflect the varying facial expressions seen in the GT. Both CombCN and LGTSM lack the ability to access sufficient information from neighboring frames, resulting in a failure to retain accurate facial identity and expressions. This misalignment can be clearly seen in the fifth and sixth row in Figure~\ref{qual_res}, where the eye gaze direction is incorrectly rendered compared to the original unmasked frames.

\subsubsection{Face geometry reconstruction}
Figure~\ref{Ablation} presents the qualitative results of our ablation study, which examines the effect of using different numbers of facial landmarks for RGB video inpainting and 3D face geometry reconstruction. The figure shows reconstructed facial frames and their corresponding 3D meshes with varying landmark configurations: 216, 68, 20, and 10.

The results demonstrate that using a higher number of landmarks (216) yields the most realistic and accurate reconstructions, both in the inpainted RGB frames and the 3D meshes. The faces reconstructed with 216 landmarks closely resemble the ground truth, capturing subtle details like facial expressions and textures, while the 3D meshes are smooth and well-defined. In fact, when visually comparing the final 3D outputs, the results for different landmark configurations appear to be very close to one another. However, when closely observed, there are differences in finer details, especially around the eye and eyebrow areas, where the model with fewer landmarks shows a slight reduction in sharpness and detail.

As the number of landmarks decreases to 68, there is a slight reduction in detail and continuity of expressions. The 3D geometry is still recognizable, but finer details, such as the curvature of the eyebrows or the contours of the eyes, are less defined. With 20 landmarks, the degradation becomes more apparent, showing reduced sharpness and increased artifacts in both the inpainted frames and the 3D meshes. The model with only 10 landmarks produces the least accurate reconstructions, with noticeable blurring and misalignment of facial features, particularly around the eye and eyebrow regions, and less convincing 3D geometry overall.

Importantly, however, the overall performance degradation from 216 landmarks to as few as 10 landmarks remains relatively small, especially when considering standard quantitative metrics (e.g., Chamfer Distance, RMS Error, Hausdorff Distance). This robustness highlights a key strength of our approach: the ability to maintain high-quality reconstruction even under sparse landmark conditions, as would occur in typical consumer-grade HMD setups.

Although the differences between the final 3D outputs are subtle, they can be observed when examining the fine details of facial features. Quantitative metrics, such as Chamfer Distance, RMS Error, and Hausdorff Distance, may show minimal differences between these configurations, but when comparing them qualitatively, the 216 landmark configuration provides a much sharper and more accurate reconstruction. This highlights that the inclusion of more landmarks allows for more precise facial geometry, especially around critical areas such as the eyes and eyebrows.

Thus, while more landmarks help refine fine details, our model remains highly effective and resilient even with significantly fewer landmarks, aligning with the practical constraints and goals of real-world XR applications.

\section{Discussion}
\label{sec:discussion}

\begin{figure}
\begin{center}
  \includegraphics[scale=0.4]{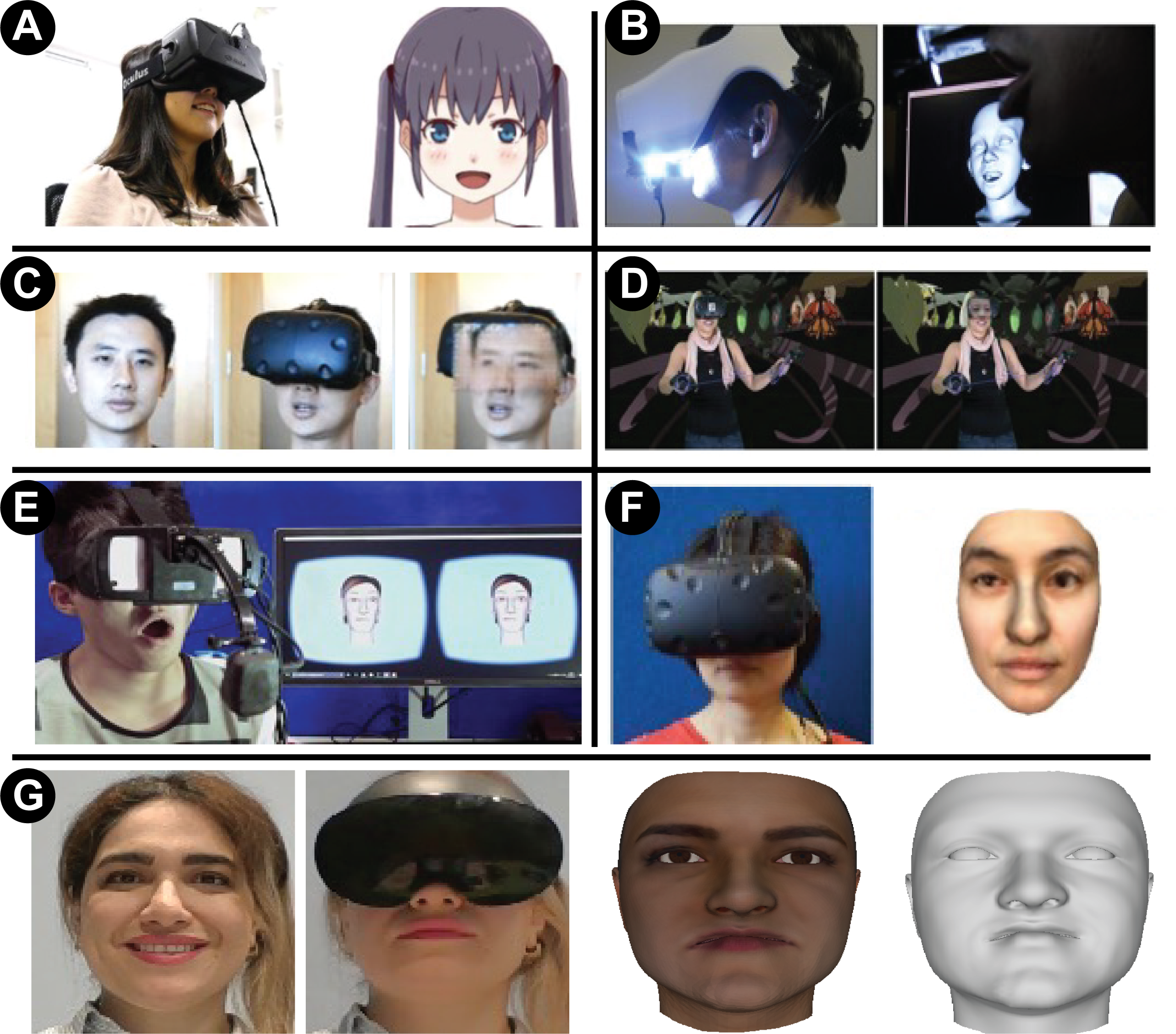}   
  \caption{Comparison of various HMD removal and facial reconstruction methods in real-world scenarios. (A) ~\cite{suzuki2017recognition}, (B) ~\cite{olszewski2016high}, (C) ~\cite{zhao2018identity}, (D) ~\cite{frueh2017headset}, (E) ~\cite{chen20223d}, (F) ~\cite{lou2019realistic}, and (G) Ours. Each method is shown in its native capture configuration; identical input images are not available across methods.
  }
    \label{real_world}
\end{center}
\end{figure}

Figure~\ref{real_world} compares various methods for HMD removal and facial reconstruction in real-world scenarios. We emphasize that Figure~\ref{real_world} is intended to illustrate the practical visual outputs of different headset-removal pipelines in their native hardware or capture environments, rather than for direct pixel-wise benchmarking. Because these methods rely on varied sensor types (e.g., RGB-D cameras, EMG sensors, monocular RGB inputs), perfectly matching input frames across approaches would be infeasible. For frame-matched and standardized quantitative comparisons, we refer readers to the benchmark evaluations presented in Section 4.

The first method, shown in Figure~\ref{real_world} (A) ~\cite{suzuki2017recognition}, uses a 2D cartoon image to represent the user’s face. Although this approach is computationally efficient, it lacks the realism necessary for an immersive experience in XR applications. It fails to capture the user’s true facial identity and expressions, which are crucial for maintaining a sense of presence in virtual environments.

In contrast, the method illustrated in Figure~\ref{real_world} (B) ~\cite{olszewski2016high} employs an external RGB/RGBD camera attached to the VR HMD to capture facial movements. While this setup allows for the tracking of expressions, it does not preserve the unique facial texture and identity of the user, leading to avatars that feel less personalized and realistic. Also, it makes the HMD more bulky. 

Figure~\ref{real_world} (C) ~\cite{zhao2018identity} showcases an image-based inpainting approach similar to ours, but without incorporating 3D modeling or temporal consistency. The absence of these elements results in cartoonish outputs with limited depth and detail, reducing the overall realism and effectiveness of the reconstructed faces.

The method depicted in Figure~\ref{real_world} (D) ~\cite{frueh2017headset} attempts to remove HMD occlusion but only does so partially, failing to fully reconstruct the face. This leads to visible artifacts that detract from the visual quality and realism of the output.

Figure~\ref{real_world} (E) ~\cite{chen20223d} on the other hand features depth camera-based methods that generate facial geometry. Although these methods can achieve accurate 3D reconstructions, they require additional hardware and complex setups, which limit their practicality for everyday use, particularly in consumer-grade XR applications.

In Figure~\ref{real_world} (F) ~\cite{lou2019realistic}, a method relying on EMG sensors is shown. This technique can only produce a limited range of facial expressions, restricting their utility in dynamic XR environments where a full spectrum of expressions is often required.

Finally, our proposed system, as illustrated in Figure~\ref{real_world} (G), stands out by reconstructing a realistic 3D face avatar from a single RGB view without the need for additional hardware. To demonstrate our approach in a real-world testing scenario, we captured data from a user wearing a Meta Quest Pro HMD with a standard webcam positioned in front of them. For testing purposes, we used a single reference frame captured when the user briefly removed the HMD, from which we detected 216 dense landmarks in the upper facial region (corresponding to the HMD-occluded area). These landmarks ideally simulate the facial tracking data that would be obtained from HMD sensors in practice. Combined with the RGB video feed and the reference frame for identity preservation, our trained model successfully generates realistic face geometry nd inpainted RGB frames. 

For visualization purposes, the reconstructed 3D meshes are textured using UV texture maps from the publicly available FFHQ-UV dataset~\cite{bai2023ffhq}. Specifically, while the 3D mesh topology and inpainted RGB outputs are generated by our model, the UV texture maps are taken from FFHQ-UV to provide appearance details during rendering. The results demonstrate high fidelity and natural expression reproduction, setting a new direction for practical XR applications.

Despite the clear advantages of our method, several limitations must be acknowledged. One significant challenge is the shadowing effect caused by the HMD on the lower part of the face, which was not addressed in this study. These shadowing effects could lead to inaccuracies in facial reconstruction, particularly under varying lighting conditions in real-world applications. Additionally, the use of synthetic masks to simulate HMD occlusions in our experiments does not capture the complexity of real-world HMD shadows and occlusions. This simplification might result in less accurate inpainting when applied to more complex, real-world scenarios.

Another critical factor is the accuracy of facial landmark detection, which is pivotal to the success of our inpainting process. In our experiments, landmark detection was applied to the ground truth data. However, in real-world applications, landmarks obscured by the HMD would need to be captured using internal cameras, which could degrade the accuracy of the detection algorithm due to factors such as image quality and the complexity of facial expressions. In scenarios where facial landmarks are not accurately detected, the inpainting process could be adversely affected, leading to less accurate facial reconstructions.

In summary, while our proposed method demonstrates significant improvements over existing techniques, addressing these limitations will be essential for enhancing its applicability and performance in real-world XR environments. Future work should focus on refining the handling of shadowing effects, improving the simulation of complex occlusions, and enhancing facial landmark detection under challenging conditions. By overcoming these challenges, our method can be further optimized to deliver even more reliable and realistic facial reconstructions, thereby advancing the state of the art in XR applications.

\section{Conclusion and future work}
\label{sec:conclusion}

This study introduces a novel learning-based framework to tackle HMD occlusion in XR applications, enhancing immersive experiences in contexts such as teleconferencing. Our method combines a GAN-based video inpainting approach with 3D Morphable Models (3DMM) and facial landmark regression to reconstruct complete 3D facial geometry from a single RGB view. The framework leverages dense landmark optimization to effectively restore missing facial information, producing photorealistic 3D face reconstructions.

Experimental results demonstrate that our framework successfully removes HMDs while retaining facial features and identity, achieving high-quality, realistic outputs. The ablation study further reveals that increasing the number of facial landmarks significantly enhances reconstruction quality and photorealism.

However, some limitations remain, such as the inability to handle shadowing effects caused by the HMD and inaccuracies during large head movements. Additionally, the use of synthetic masks may not fully capture real-world occlusion complexity. Future work will address these issues by refining the handling of shadowing effects and improving performance during large head movements. We aim to enhance face geometry consistency, enable real-time 3D reconstruction, and synchronize with audio to improve the applicability of our method in dynamic XR environments.

In addition, future research will systematically benchmark the trade-offs between computational complexity and reconstruction quality. A broader evaluation across demographically diverse human faces and expression analysis will also be conducted to improve model generalization. Furthermore, a multi-view calibrated setup with HMD occlusions will be established to enable evaluation of side-view synthesis fidelity, aligning more closely with real-world XR deployment scenarios.

\subsection*{Data Availability Statement}
The data supporting the findings of this study will be made available upon reasonable request. Researchers may contact the first author for access to datasets and further information.

\subsection* {Acknowledgments}
This project has received funding from the European Union’s Horizon 2020 research and innovation program under the Marie Skłodowska-Curie grant agreement No 956770.
\subsection* {Code, Data, and Materials Availability}
The implementation code for this study will be made publicly available on GitHub. Researchers are encouraged to contact the first author for further guidance or inquiries.


\bibliography{report}   
\bibliographystyle{spiejour}   


\vspace{2ex}\noindent\textbf{Fatemeh Ghorbani Lohesara} is currently a Marie Curie Fellow and a Ph.D. student in the Communication Systems Group at Technische Universität Berlin, Germany. She received her M.Sc. degree in mechatronics from K. N. Toosi University, and holds a B.Sc. in electrical engineering from Guilan University. Her research focuses on addressing the headset removal problem for gaze contact in XR applications. Her main research interests include human–computer interaction, XR, and serious games.

\noindent\textbf{Karen Eguiazarian (Egiazarian)} (Fellow,
IEEE) (SM’96, F’18) received the M.Sc. degree in mathematics from Yerevan State University, Armenia, in 1981, the Ph.D. degree in physics and mathematics from Moscow State University, Russia, in 1986, the Doctor of Technology degree from the Tampere University of Technology (TUT), Tampere, Finland, in 1994. He is a Professor with the Signal Processing Department, Tampere University, leading the Computational Imaging Group, and a Docent with the Department of Information Technology, University of Jyväskylä, Finland. His main research interests are in the fields of computational imaging, compressed sensing, efficient signal processing algorithms, image/video restoration, and image compression. He has published over 750 refereed journal and conference articles, books, and patents in these fields. Prof. Egiazarian is a member of the DSP Technical Committee of the IEEE Circuits and Systems Society. He has served as an associate editor in major journals in the field of his expertise, including the IEEE Transactions on Image Processing, and was Editor-in-chief of the Journal of Electronic Imaging (SPIE).

\noindent\textbf {Sebastian Knorr} (SM'19, IEEE) received the Dipl.-Eng. and Dr.-Eng. degree in electrical engineering from Technical University of Berlin in 2002 and 2008, respectively. He is professor at the HTW Berlin. His main research interests are in the field of computer vision, 3D image processing and immersive media, in particular virtual reality applications. 
Prof. Knorr received the German Multimedia Business Award of the Federal Ministry of Economics and Technology in 2008, and was awarded by the initiative “Germany-Land of Ideas” which is sponsored by the German government, commerce and industry in 2009, respectively. He received a couple of best paper awards including the Scott Helt Memorial Award of the IEEE Transactions on Broadcasting in 2011 and the Lumiére Award at the International Conference on 3D Immersion in 2018. Prof. Knorr has served as an associate editor for the IEEE Trans. on Multimedia and is currently serving as an associate editor for the IEEE Trans. on Image Processing. 
\newline

\vspace{1ex}

\listoffigures
\listoftables

\end{spacing}
\end{document}